\documentclass[11pt]{article}
\usepackage{acl2016}
\usepackage{times}
\usepackage{url}
\usepackage{latexsym}

\usepackage[cmex10]{amsmath}
\usepackage{amssymb}
\usepackage{amsthm}
\usepackage{bauer}

 \usepackage[pdftex]{graphicx}
\graphicspath{{figures/}}

\aclfinalcopy 

\title{Optimizing for Measure of Performance in Max-Margin Parsing}

\author{
  Alexander Bauer$^{1,2}$, \hspace*{5pt}Shinichi Nakajima$^{1,2}$,\hspace*{5pt} Nico G{\"o}rnitz$^{2}$,\hspace*{5pt} Klaus-Robert~M{\"u}ller$^{1,2,3,4}$\\
  $^1$Berlin Big Data Center, Berlin, Germany\\
  $^2$Machine Learning Group, Technische Universit{\"a}t Berlin, Berlin, Germany\\
  $^3$Max Planck Institute for Informatics, Saarbr{\"u}cken, Germany\\
  $^4$Department of Brain and Cognitive Engineering, Korea University, Seoul, Korea\\
  \texttt{\{alexander.bauer, nakajima, klaus-robert.mueller\}@tu-berlin.de} \\
}

\date{}

\begin{document}
\maketitle
\begin{abstract}
Many statistical learning problems in the area of natural language processing
including sequence tagging, sequence segmentation and syntactic parsing
has been successfully approached by means of structured prediction methods.
An appealing property of the corresponding discriminative learning algorithms
is their ability to integrate the loss function of interest directly into the
optimization process, which potentially can increase the resulting performance accuracy.
Here, we demonstrate on  the example of constituency parsing how to optimize for $F_1$-score
in the max-margin framework of structural SVM.
In particular, the optimization is with respect to the original (not binarized) trees.
\end{abstract}

\section{Introduction}
\label{intro}

Many statistical learning problems in the area of natural language processing (NLP)
including sequence tagging, sequence segmentation and various kinds of syntactic parsing
have been successfully approached by means of structured prediction methods, which
correspond to a machine learning paradigm that considers learning with complex outputs
like sequences, trees or even general graphs.
Popular examples of the corresponding methods include
maximum margin Markov networks (M$^3$N) \cite{Taskar03max-marginmarkov}, structural support vector machine (SSVM) \cite{Tsochantaridis05largemargin},
and on-line algorithms like MIRA \cite{McDonaldCP05}.

Apart from maximizing the margin between the true and false outputs, another
appealing property of these discriminative learning algorithms is their ability
to incorporate the loss function of interest directly
in the training procedure, which potentially
can improve the resulting prediction accuracy.
However, the existing training approaches
including cutting-plane algorithm \cite{Joachims09lcuttingPlane}, bundle methods \cite{SmolaVL07} and 
Frank-Wolfe optimization \cite{Lacoste-JulienJSP13} assume that an efficient inference algorithm
is given during the training in order to compute a subgradient
of the objective function or the most violating output
with respect to a given loss function.
This usually results in a combinatorial problem which often can be
solved by means of dynamic programming.
The success of the latter crucially depends
on the form of the underlying model and the chosen loss function.
Usually, if both decompose over small sets of variables
we can apply efficient inference algorithms e.g. Viterbi algorithm \cite{Forney73} for
sequence tagging, CKY algorithm \cite{Younger1967189} for syntactic parsing
or sum-product belief propagation \cite{Bishop:2006:PRM:1162264} for probabilistic inference.
Still, some popular performance measure like precision in information retrieval
or $F_1$-score in segmentation and parsing tasks
do not decompose in this way and, therefore, are often referred to as 
high-order measures.
Nevertheless, for non-decomposable loss functions which are build by a composition
of locally decomposable statistics and some non-decomposable
wrapper function, we can perform inference
efficiently in polynomial time as has been shown in \cite{auli-lopez:2011:EMNLP} and   \cite{Bauer2016}.

In this paper we consider only the task of syntactic parsing,
an important preprocessing step for many NLP applications
which aim at processing the meaning of a natural text.
In particular,  we focus
on the constituency parsing \cite{Taskar04max-marginparsing}, \cite{Bauer2015}  where the goal is, for a given input sentence, to predict the
most probable parse tree according to some context-free grammar.
A lot of progress has been done previously
in order to train a corresponding model based on finite-state techniques like probabilistic context free grammars (PCFGs) \cite{Johnson98pcfgmodels}
or more general weighted context free grammars (WCFGs) \cite{Tsochantaridis05largemargin}.
Here we build on the discriminative max-margin approach of SSVM
aiming at optimizing $F_1$-score with respect to the constituents of parse trees.

In order to achieve a cubic running time for prediction it is a conventional approach to binarize the grammar before training.
Unfortunately, this introduces a bias during the training procedure as the corresponding loss function is evaluated on the binary representation
while the resulting performance is measured on the original not binarezd trees.
In this paper we extend the inference procedure presented in \cite{Bauer2016} to account for this difference. The result corresponds to the inference on not binarized trees
leading to a better prediction accuracy while keeping the computational advantage of binarized representation.



\section{Optimizing for $F_1$-score}
\label{section2}
A common structured prediction approach is to learn
a functional relationship $f \colon \mcX \rightarrow \mcY$
between an input space $\mcX$ and an arbitrary discrete output space $\mcY$ of the form
\begin{equation}
\label{E_prediction}
f(x) = \underset{y \in \mcY}{\argmax} \hspace*{5pt}w^\T \Psi(x, y).
\end{equation}
Here, $\Psi \colon \mcX \times \mcY \rightarrow \mathbb{R}^d$
is the joint feature map describing the compatibility between an input $x$
and a corresponding output $y$, and $w$ is the vector of model weights to be learned
from a training sample of input-output pairs $(x_1, y_1), ..., (x_n, y_n) \in \mcX \times \mcY$.
Training the models weights $w$ according to the maximum-margin criterion
of an SSVM corresponds to solving the following
optimization problem
\begin{equation}
\begin{aligned}
\underset{w,\hspace*{2pt} \xi \geqslant 0}{\text{min.}} & \hspace*{5pt} \frac{1}{2}\|w\|^2 + \frac{C}{n}\sum_{i = 1}^n \xi_i  \hspace*{10pt}  \text{subject to}\\
& w^\T \left(\Psi(x_i, y_i^*) - \Psi(x_i, y)\right) \geqslant 1   - \frac{\xi_i}{\Delta(y^*_i, y)}\\
& \forall i \in \{1, ..., n\}, \hspace*{1pt}\forall y \in \mcY\backslash \{y_i^*\}
\end{aligned}
\label{E_margin_scaling}
\end{equation}where
$C$ is a regularization constant and  
$\Delta \colon \mcY \times \mcY \rightarrow \mathbb{R}$
denotes a corresponding loss function quantifying the discrepancy
between a prediction $y$ and the ground truth $y^*$.

As already mentioned in the introduction
the most popular training  approaches
rely on the assumption that during training process we are given an additional
inference algorithm to compute the subgradient (in case of bundle methods)
or the most violating configuration (in case of cutting-plane approach).
In the literature, this problem
is referred to as the \emph{loss augmented} inference \cite{TaskarCKG05}, \cite{Bauer2013}.
In the most general form it corresponds to maximizing
the following objective
\begin{equation}\label{E_LAI}
\underset{y \in \mcY}{\max}\hspace*{5pt}\Delta(y^*, y) \cdot (\text{const.} + w^\T \Psi(x,y))
\end{equation}
where the constant term is  $1 - w^\T \Psi(x,y^*)$.
The main computational difficulty here arises from the fact
that the size of $\mcY$ may grow exponentially in the size of the input $x$
as it is the case in constituency parsing.
Therefore, in order to perform inference efficiently
we have to restrict the range of possible models and loss functions.
An important observation here is that $F_1$-score can be parameterized by the
number of true (TP) and false (FP) positives
according to
\begin{equation}
F_1(TP, FP) = \frac{2TP}{|y^*|+TP+FP}
\end{equation}
where $|y^*|$ denotes the number of nodes in the true parse tree.
In particular, these counts (TP and FP) decompose over
the individual nodes of parse trees.
The main idea for solving (\ref{E_LAI}) is then to stratify the maximisation
over all configurations of (TP, FP) and picking the best.

\subsection{Model Description}
For the task of constituency parsing the set $\mcY$ in (\ref{E_prediction}) corresponds to
all valid parse trees with respect to the given grammar $G$ and an input sentence $x$.
A popular approach for representing a parse tree $y \in \mcY$
is based on a joint feature vector $\Psi(x,y)$ where the individual dimensions correspond
to the grammar rules and the entries are the counts how often a production rule from $G$
occurs in a tree $y$. The dimension of the resulting feature vector is therefore
equal to the number of different production rules in the grammar.
Furthermore, due to such representation the score $w^\T \Psi(x,y)$
for each pair $x,y$ decomposes over individual productions in the tree $y$
enabling efficient dynamic programming algorithms (e.g., CKY algorithm).

\subsection{Grammar Binarization}
In order to achieve a cubic running time (in the length of the sentence)
it is a common approach to binarize a given grammar (or equivalently the trees) before training
by the left or right factorization introducing new artificial constituents
as illustrated in Figure 1.
We can vary the amount of annotation (e.g., number of missing nodes to the right in case of right-factorization) contained in these artificial
constituents, which is referred to as the horizontal annotation.
Another useful annotation technique in order to increase the expressivity of the grammar
is the parent annotation where the labels of individual nodes in a parse tree are extended
by the label of their parent nodes introducing more contextual information into the labels
(see Figure 1).

\begin{figure}[t]
\label{bauer1}
\centering
\includegraphics[scale = 0.6]{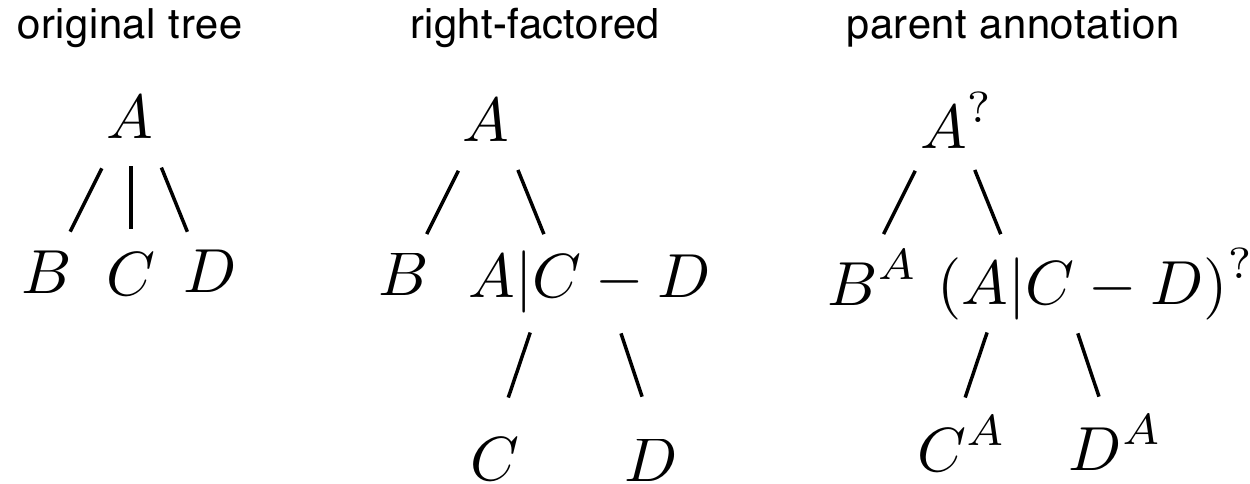}
\caption{Grammar binarization due to binarization of trees.
In the notation of artificial constituents $A|C-D$, $A$ before $|$
denotes the parent in the original tree and $C-D$ the children nodes
of $A$
which are spanned by the current artificial constituent. $"?"$
denotes the parent label of $A$.
}
\end{figure}

\subsection{Loss Augmented Inference}
We now show how the problem in (\ref{E_LAI})
can be solved via dynamic programming for 
constituency parsing with $\Delta_{F_1}(y^*,y) = 1 - F_1(y^*,y)$.
Let an input sentence $x = (x_1, ..., x_{|x|})$ with $x_i$ denoting the token on position $i$ and a corresponding
true parse tree $y^*$ be given.
Similar to the conventional CKY algorithm the idea here is to
iteratively compute the values for the subproblems
\begin{equation}\label{E_Pi_L}
\Pi_{i,j,A}^{tp,fp} := \underset{y \in \mathcal{T}_{i,j,A}^{tp,fp}}{\max}\hspace*{5pt}\sum_{p \in \text{Productions}(y)} q(p),
\end{equation}where $\mathcal{T}_{i,j,A}^{tp,fp}$ denotes the set of all valid subtrees spanning the tokens $(x_i, ..., x_j)$ and having the label $A$ at the root.
The parameters $tp$, $fp$ encode the number of true and false positives with respect to $y^*$.
$q(p)$ denotes the weight of a production $p$ times its frequency in the parse tree $y$.
That is, the quantity $\Pi_{i,j,A}^{tp,fp}$ denotes the value 
of an optimal label configuration over the parse trees
$y \in \mathcal{T}_{i,j,A}$
which additionally result in a fixed value
for true and false positives.
The values of these subproblems can be computed 
in a bottom-up manner according to the following equation
\begin{eqnarray}
\label{E_inference}
\Pi_{i,j,A}^{tp,fp} = &\underset{A \rightarrow B \hspace*{1pt}C,\hspace*{2pt} s,\hspace*{2pt} \hat{tp},\hat{fp}}{\max}\hspace*{5pt}& q(A \rightarrow B C) + \nonumber \\
&&\Pi_{i,s,B}^{\bar{tp}, \hspace*{2pt}\bar{fp}} + \nonumber \\
&&\Pi_{s+1, j, C}^{\hat{tp}, \hat{fp}}
\end{eqnarray}
where we maximize over all possible grammar productions $A \rightarrow B \hspace*{1pt}C$ with fixed $A$, over all split points of subtrees $i \leqslant s < j$, and over possible distributions of the loss parameters
$\hat{tp}$ and $\hat{fp}$ with $\bar{tp} := tp - \mathbf{1}([A]_{ij} \hspace*{1pt} \in \hspace*{1pt}y^*) - \hat{tp}$ and $\bar{fp} := fp - \mathbf{1}([A]_{ij} \hspace*{1pt} \notin \hspace*{1pt}y^*) - \hat{fp}$. The term $[A]_{ij}$ denotes
a node in a subtree that spans tokens $x_i, ..., x_j$ and has the label $A$,
and $\mathbf{1}(\cdot)$ is an indicator function yielding $1$ if the expression inside the brackets is true
and $0$ otherwise.
With a slight abuse of notation we write $[A]_{ij} \in y^*$ to check if a node with the corresponding label is in a tree $y^*$.

After computing the values for all the subproblems,
we can obtain 
the optimal value $p^*$ of the problem in (\ref{E_LAI})
by maximizing over all possible values $tp, fp$ according to
\begin{equation}\label{E_termination}
p^* = \underset{tp, fp}{\max}\hspace*{2pt} (1 - F_1(tp, fp)) \cdot (const. + \Pi_{1,|x|,\mathcal{S}}^{tp, fp}).
\end{equation}
where $\text{const.}$ corresponds to the constant term $1 - w^\T \Psi(x,y^*)$.
$|x|$ denotes the number of tokens in the input sentence
and $\mcS$ is the start (or root) symbol of each parse derivation.
The corresponding maximizing argument can be found by backtracking the optimal
decisions in each computation step as usually done
in dynamic programming.

Note, that the counts of true and false positives in the above computation scheme
is with respect to the binarized tree representation.
The resulting performance, however, is evaluated on the
original tree representation after reversing the binarization.
It turns out that we can easily adjust the above computation scheme
to keep track of the corresponding counts with respect to unbinarized trees.
First note that in order to transform a binarized tree in the original form we need to remove
all the artificial constituents, that is the counts of true and false positives are not affected
by their presence. Furthermore, after removing an artificial constituents we need to attach its
children in a tree to its parent. In particular, the boundary indices of the corresponding spans of
the children nodes do not change during this procedure.
Finally we have to remove the additional annotation from the labels of the remaining nodes.
To summarize,
we can compute the counts of true and false positives with respect to unbinarized grammar
from binarized trees if we completely ignore artificial nodes and the additional annotation (e.g. parent annotation).
More precisely, we only need to replace the indicator function $\mathbf{1}$ for computing $\bar{tp}, \bar{fp}$ in (\ref{E_inference})
by
\begin{equation}
\bar{\mathbf{1}}([A^?]_{ij} \in y^*)=\begin{cases}
  0,  & A \text{ is artificial,}\\
  \mathbf{1}([A]_{ij} \in y^*), & \text{else}
\end{cases}
\end{equation}
where $?$ denotes the parent annotation of $[A]_{ij}$ and $y^*$ corresponds to the (unbinarized) ground truth.
Similarly, we define 
\begin{equation}
\bar{\mathbf{1}}([A^?]_{ij} \notin y^*)=\begin{cases}
  0,  & A \text{ is artificial,}\\
  \mathbf{1}([A]_{ij} \notin y^*), & \text{else}
\end{cases}
\end{equation}

This way we ensure that the corresponding counts of true and false positives
are with respect to the unbinarized trees. The overall
computation scheme is provable correct, that is
it computes the optimal value of the problem in ($\ref{E_LAI}$).

\section{Experiments}
\label{section3}

In this section we present our preliminary experimental results
for the task of constituency parsing by training an SSVM
via cutting plane algorithm and optimizing for $F_1$-score.
In particular, we compare the performance when optimizing on trees in the binarized representation
(marked by "(bin.)") versus  the original non binarized trees.
Additionally we report results when optimizing for 0/1 accuracy, and
the number of false positives (\#FP).
The resulting perforce is evaluated in terms of precision (P), recall (R), 
$F_1$-score ($F_1$), and 0/1 prediction accuracy (all with respect to the unbinarized trees).

\begin{figure}[t]
\centering
\includegraphics[scale = 0.38]{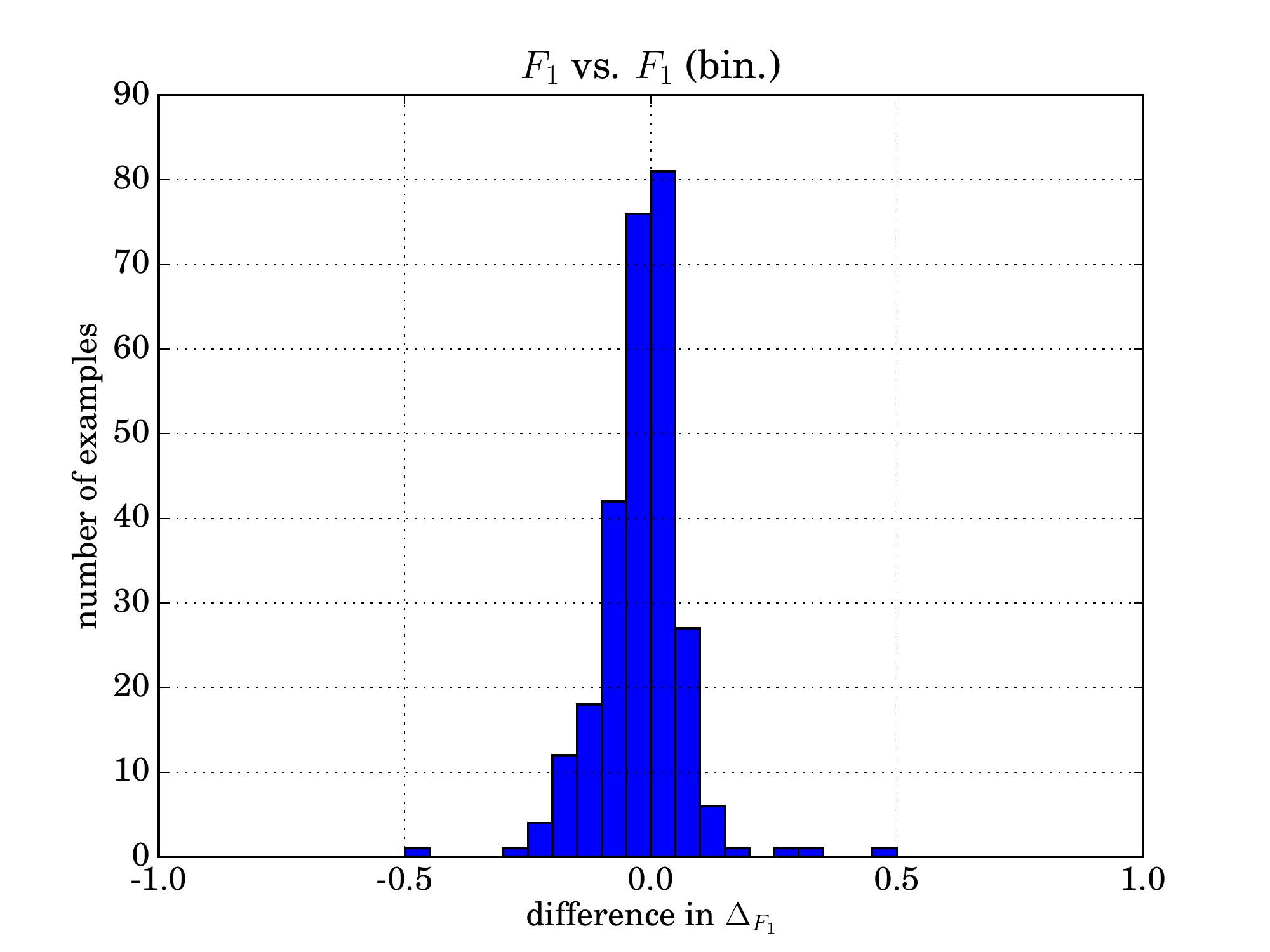}
\caption[textformat=period]{Illustration of  the loss difference (in $\Delta_{F_1}$) on the test set when optimizing for $F_1$ score on the original trees 
versus binary representation. 
}
\label{fig_wilcoxon}
\end{figure}

As training data we used a subset of the Wall Street Journal (WSJ) from
the Penn English Treebank-3 restricted to sentences of the length $\leqslant$ 20.
We used the standard data split: sections 2-21 of WSJ for training (16667 sentences),
section 22 for validation (725 sentences) and section 23 for testing (1034 sentences).
The parse trees were preprocessed in the standard way by removing functional
tags and null elements. The regularization hyperparameter $C$
was chosen by cross-validation over a grid of values $\{10^i \colon i = 0,1, ...5\}$.
We report the corresponding results on the test data in Table \ref{table2}.
The first column describes the measure we optimized during the training procedure.
Here we can make two observations.
First, we see that there is a little difference in performance between \#FP (bin.) and $F_1$-score (bin.)
on binarized trees supporting the claims in \cite{Bauer2015} (see Proposition 2).
Second, we see that adjusting training with $F_1$-score for unbinarized trees
improves the resulting performance upon training with binarized representation.
According to the Wilcoxon signed-rank test \cite{ReyN11} this result is statically significant.
Figure \ref{fig_wilcoxon} illustrates the difference in loss $\Delta_{F_1}$ for models optimized according to original versus binarized representation.
A corresponding null-hypothesis is that the measurement difference for a pair of methods 
follows a symmetric distribution around zero. Here, we can see a clear shift of the corresponding distribution to the left\footnote{Note that we do not count examples with zero difference in the loss value.}.


\begin{table}[t]
\caption{Experimental results on the test set.}
\label{table2}
\begin{center}
\begin{tabular}{|l|c|c|c|c|}
\hline
Measure & $\textit{P}$ & $\textit{R}$ & $F_1$ & 0/1 Acc  \\ \hline \hline
$\textrm{0/1}$ (bin.) & $88.86$ & $89.27$ & $89.00$& $27.32$\\
$\textit{\#FP}$ (bin.) & $90.43$ & $90.34$ & $90.32$ & $27.91$\\
$F_1$ (bin.) & $90.33$ & $90.43$ & $90.32$ & $28.69$\\
$F_1$ & $\boldsymbol{90.92}$ & $\boldsymbol{90.82}$ & $\boldsymbol{90.82}$ & $\boldsymbol{29.18}$\\

\hline
\end{tabular}
\end{center}
\end{table}

\section{Conclusion}
\label{section4}
We demonstrated on the example of constituency parsing
how to optimize the weights of the model with respect to $F_1$-score
in the maximum-margin framework of SSVMs.
In particular, we showed how the optimization during the training procedure can be performed with respect to the original non binarized trees.
More precisely,
the proposed modification allows 
to perform loss augmented inference on non binarized trees, which results in a better prediction accuracy, while keeping the computational advantage of binarized representation.
Our preliminary experimental results suggest an improvement in the prediction performance
by applying this new technique.
According to the Wilcoxon signed-rank test the presented performance difference is statically significant.


\subsubsection*{Acknowledgments}
This work was supported by the Federal Ministry of Education and Research under the Berlin Big Data Center Project
under Grant FKZ 01IS14013A. The work of K.-R. M{\"u}ller was supported in part by the BK21 Program of NRF Korea,
BMBF, under Grant 01IS14013A and by Institute for Information and Communications Technology Promotion (IITP) grant funded by the Korea government (No. 2017-0-00451)

\bibliography{acl2016}

\begin{thebibliography}{}

\bibitem[\protect\citename{Auli and Lopez}2011]{auli-lopez:2011:EMNLP}
Michael Auli and Adam Lopez.
\newblock 2011.
\newblock Training a log-linear parser with loss functions via softmax-margin.
\newblock In {\em Proceedings of the 2011 Conference on Empirical Methods in
  Natural Language Processing}, pages 333--343, Edinburgh, Scotland, UK., July.
  Association for Computational Linguistics.

\bibitem[\protect\citename{Bauer \bgroup et al.\egroup }2013]{Bauer2013}
Alexander Bauer, Nico G\"ornitz, Franziska Biegler, Klaus-Robert M\"uller, and
  Marius Kloft.
\newblock 2013.
\newblock Efficient algorithms for exact inference in sequence labeling svms.
\newblock {\em IEEE Trans. on Neural Netw. and Learning Syst.}, 25(5):870--881,
  Oct.

\bibitem[\protect\citename{Bauer \bgroup et al.\egroup }2016]{Bauer2016}
Alexander Bauer, Shinchi Nakajima, and Klaus-Robert M\"uller.
\newblock 2016.
\newblock Efficient exact inference with loss augmented objective in structured
  learning.
\newblock {\em IEEE Trans. on Neural Netw. and Learning Syst.; In Print}.

\bibitem[\protect\citename{Bauer \bgroup et al.\egroup }2017]{Bauer2015}
Alexander Bauer, Mikio Braun, and Klaus-Robert M\"uller.
\newblock 2017.
\newblock Accurate maximum-margin training for parsing with context-free
  grammars.
\newblock {\em IEEE Trans. on Neural Netw. and Learning Syst.}, 28(1):44--56.

\bibitem[\protect\citename{Bishop}2006]{Bishop:2006:PRM:1162264}
Christopher~M. Bishop.
\newblock 2006.
\newblock {\em Pattern Recognition and Machine Learning (Information Science
  and Statistics)}.
\newblock Springer-Verlag New York, Inc., Secaucus, NJ, USA.

\bibitem[\protect\citename{Forney}1973]{Forney73}
George~David Forney.
\newblock 1973.
\newblock The viterbi algorithm.
\newblock In {\em Proc IEEE}, volume~61, pages 268--278.

\bibitem[\protect\citename{Joachims \bgroup et al.\egroup
  }2009]{Joachims09lcuttingPlane}
T.~Joachims, T.~Finley, and Chun-Nam Yu.
\newblock 2009.
\newblock Cutting-plane training of structural svms.
\newblock {\em Machine Learning}, 77(1):27--59, Oct.

\bibitem[\protect\citename{Johnson}1998]{Johnson98pcfgmodels}
Mark Johnson.
\newblock 1998.
\newblock {PCFG} models of linguistic tree representations.
\newblock {\em Computational Linguistics}, 24(4):613--632.

\bibitem[\protect\citename{Lacoste{-}Julien \bgroup et al.\egroup
  }2013]{Lacoste-JulienJSP13}
Simon Lacoste{-}Julien, Martin Jaggi, Mark~W. Schmidt, and Patrick Pletscher.
\newblock 2013.
\newblock Block-coordinate frank-wolfe optimization for structural svms.
\newblock In {\em Proc. 30th ICML}, pages 53--61, Jun.

\bibitem[\protect\citename{McDonald \bgroup et al.\egroup }2005]{McDonaldCP05}
Ryan~T. McDonald, Koby Crammer, and Fernando C.~N. Pereira.
\newblock 2005.
\newblock Online large-margin training of dependency parsers.
\newblock In {\em {ACL} 2005, 43rd Annual Meeting of the Association for
  Computational Linguistics, Proceedings of the Conference, 25-30 June 2005,
  University of Michigan, {USA}}.

\bibitem[\protect\citename{Rey and Neuh{\"{a}}user}2011]{ReyN11}
Denise Rey and Markus Neuh{\"{a}}user.
\newblock 2011.
\newblock Wilcoxon-signed-rank test.
\newblock In {\em International Encyclopedia of Statistical Science}, pages
  1658--1659. Springer.

\bibitem[\protect\citename{Smola \bgroup et al.\egroup }2007]{SmolaVL07}
Alexander~J. Smola, S.~V.~N. Vishwanathan, and Quoc~V. Le.
\newblock 2007.
\newblock Bundle methods for machine learning.
\newblock In {\em Proc. 21st NIPS}, pages 1377--1384, Vancouver, British
  Columbia, Canada, Dec.

\bibitem[\protect\citename{Taskar \bgroup et al.\egroup
  }2003]{Taskar03max-marginmarkov}
Benjamin Taskar, Carlos Guestrin, and Daphne Koller.
\newblock 2003.
\newblock Max-margin markov networks.
\newblock In {\em Proc. 16th NIPS}, pages 25--32, Dec.

\bibitem[\protect\citename{Taskar \bgroup et al.\egroup
  }2004]{Taskar04max-marginparsing}
Ben Taskar, Dan Klein, Michael Collins, Daphne Koller, and Christopher~D.
  Manning.
\newblock 2004.
\newblock Max-margin parsing.
\newblock In {\em Proc. EMNLP}, pages 1--8, Barcelona, Spain, Jul.

\bibitem[\protect\citename{Taskar \bgroup et al.\egroup }2005]{TaskarCKG05}
Benjamin Taskar, Vassil Chatalbashev, Daphne Koller, and Carlos Guestrin.
\newblock 2005.
\newblock Learning structured prediction models: a large margin approach.
\newblock In {\em Machine Learning, Proceedings of the Twenty-Second
  International Conference {(ICML} 2005), Bonn, Germany, August 7-11, 2005},
  pages 896--903.

\bibitem[\protect\citename{Tsochantaridis \bgroup et al.\egroup
  }2005]{Tsochantaridis05largemargin}
I.~Tsochantaridis, T.~Joachims, T.~Hofmann, and Y.~Altun.
\newblock 2005.
\newblock Large margin methods for structured and interdependent output
  variables.
\newblock {\em Journal of Machine Learning Research}, 6:1453--1484, Sep.

\bibitem[\protect\citename{Younger}1967]{Younger1967189}
Daniel~H. Younger.
\newblock 1967.
\newblock Recognition and parsing of context-free languages in time n{\^{}}3.
\newblock {\em Information and Control}, 10(2):189--208.

\end{thebibliography}
\bibliographystyle{acl2016}

\end{document}